# Design, Manufacturing, and Controls of a Prismatic Quadruped Robot: PRISMA

Bhavya Giri Goswami[1], Aman Verma[1], Gautam Jha[2], Vandan Gajjar[2], Vedant Neekhra[2], Utkarsh Deepak[2], Aayush Singh Chauhan[3]

Team Robocon, Indian Institute of Technology, Roorkee, Uttarakhand, India
e-mail: ( bgoswami@me, averma1@me, gjha@me, gvandan@me, vneekhra@ce, udeepak@ph, asingh@bt).iitr.ac.in

**ABSTRACT** - **Most of the quadrupeds developed are highly actuated, and their control is hence quite cumbersome. They need advanced electronics equipment to solve convoluted inverse kinematic equations continuously. In addition, they demand special and costly sensors to autonomously navigate through the environment as traditional distance sensors usually fail because of the continuous perturbation due to the motion of the robot. Another challenge is maintaining the continuous dynamic stability of the robot while walking, which requires complicated and state-of-the-art control algorithms. This paper presents a thorough description of the hardware design and control architecture of our in-house prismatic joint quadruped robot called the PRISMA. We aim to forge a robust and kinematically stable quadruped robot that can use elementary control algorithms and utilize conventional sensors to navigate an unknown environment. We discuss the benefits and limitations of the robot in terms of its motion, different foot trajectories, manufacturability, and controls.**

## I. INTRODUCTION

Today, most industrial mobile robots are wheeled robots. They are stable and can rush with high precision. However, their motions are only limited to rigid flat surfaces. There are multiple situations when wheeled robots' motions are restricted either by uneven terrain (like stairs or slopes) or by certain obstacles in the way. To overcome such a situation, legged robots, specifically quadruped robots, have increased significantly in industries.

In the current state of research in quadrupedal locomotion, the most common type of quadrupeds around us are either fully actuated (with three rotational degrees of freedom in each leg- Hip flexion, knee flexion & hip abduction) like Stoch [1], Spot mini [2], MIT Cheetah-3 [3], StarlETH [4], ANYmal [5] or under-actuated with novel leg linkage structures like Cheetaroid II [6], Minitaur [7], Stanford's Doggo [8]. Despite their impressive results and exceptional bio-mimicry, it is essential to note that they required highly complex control algorithms to maintain reliability and consistency, which are effective with advanced actuators like customized direct drive brushless motors [9], [10] or Series elastic actuators (SEA)[11]. Such actuators and advanced electronics components with customized manufacturing significantly increased their cost to a minimum of $30,000 [1]. Moreover, the torso's swaying motion due to the robot's gait makes it challenging to use economic and conventional distance sensors like ultrasonic sensors for even fundamental autonomous navigation and obstacle detection. Therefore these robots generally incorporate sophisticated sensors like cameras, 3D Lidars, or Kinect sensors for navigation, which increase both the computational power requirement and the cost of the robot.

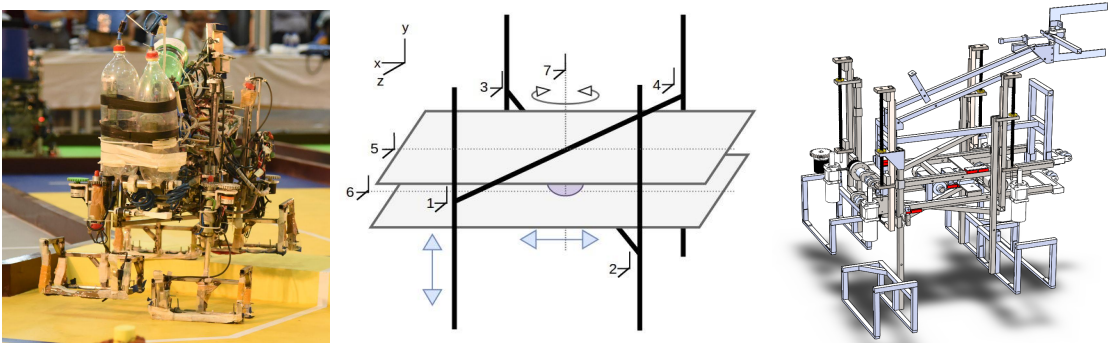

**Fig.1** - (Left) - Final Prototype, (Middle) - Simple schematic model, (Right) Detailed CAD model of PRISMA

Hence, we developed a quadruped robot with a unique prismatic design called the PRISMA (see Figure 1) to realize a robust, economical, and easy to control walking platform. This document describes this robot's design, hardware, and controlling aspects. The name of the robot is an abbreviation for PRISmatic Multi-terrain Autonomous quadruped robot. As the name suggests, it uses mainly prismatic joints for the orthogonal motion of the legs. Each leg's walking trajectory is developed by integrating linear movement parallel to the ground (horizontal) and normal to the ground (vertical) for each leg. A three-axis 3D printer inspires the movements of this robot. Prisma is a novel design idea with one vertical degree of freedom (DOF) on each leg, two DOF for the horizontal motion of the body, and one DOF for turning (for yaw), resulting in 7 DOF in the robot.

As the legs are designed to move independently in both horizontal and vertical directions, combining these motions allows the robot to make customized walking trajectories & move. Diagonal pairs of legs are joined via a sliding module to reduce the number of actuators and the robot's weight, thus limiting it only to trot gait. Trot gait [12] is well-suited for rough terrain and traveling long distances at a fair rate of speed as work is spread evenly over all four limbs. Because of diagonal support, the Zero Moment Point (ZMP) [13], [14] always remains aligned with the robot's center of mass inside the support polygon, thus maintaining the kinematic stability at every instance during walking. In addition, both the layers have an intermediary steering mechanism for the robot's turning. This ingenious steering mechanism allows Prisma to move in any direction by changing its orientation without changing its position.

Each leg has a modular design, fabricated using readily available off-the-shelf materials like Stainless Steel (SS) square tubes and custom PLA 3D printed parts. The welded SS structure is conducive for its robustness, stability, repeatability, consistency as it averts any bending or deformation. Although the robot is only limited to trot gait, it can still change speed within a range and climb obstacles by changing the stride parameters, and thus the foot trajectory. The continuous kinematic stability and robust design ensure minimal angular & vertical perturbations while walking. Therefore, SLAM-based navigations can be easily incorporated using conventional distance sensors like ultrasonic or Lidar sensors. Thus, reducing the necessity of high computational algorithms, costly sensors, and complex navigation strategies like computer vision. In addition, Prisma is adaptable to the environment, i.e., it switches its foot trajectory as per the size of the obstacle ahead. In the design stage of the robot, we have focused on the proper functioning and explored the possibility of its economic aspects and easy-to-control algorithms so that quadruped could be accessible to almost anyone, and its reproduction is feasible.

This paper is organized to illustrate the design and hardware with a detailed CAD in Section-2, Electronics, embedded system, actuators, sensor inputs, and communication interfaces in Section-3. Section-4 expounds on the Pure pursuit control algorithm variation used with all the different trajectories and their simulation. Finally, Section-5 describes the experimental result of Prisma demonstrating its different motions and its autonomous movement on variable terrain.

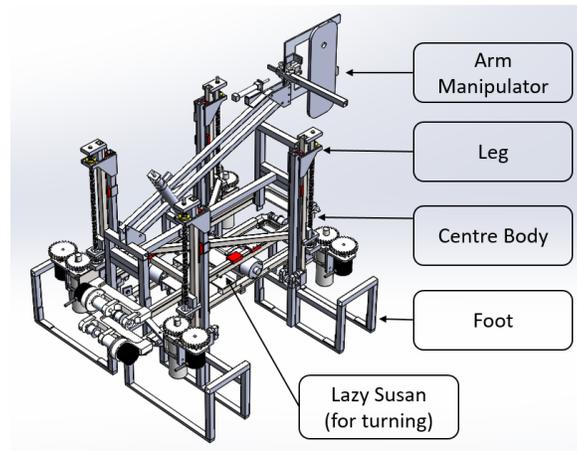

**Fig. 2 -** Schematic CAD model of Prisma

## II. MECHANICAL DESIGN

*Prisma* is a quadrupedal robot developed in-house from scratch at the Indian Institute of Technology (IIT), Roorkee, India. The design philosophy of this robot is based on robustness, cost-friendly fabrication, high body stability, ease to control, and capacity to incorporate simple autonomous navigation strategies.

The robot design can be well understood as an assembly of the central body which consists of two parallel layers with a sliding module mounted to each layer and a steering module sandwiched between them. Each sliding module connects one diagonal pair of prismatic legs and slides horizontally on the upper and lower layer. Figure-2 shows the CAD showing all the modules of Prisma. Multiple 3D printed PLA connectors and mountings were used as joints and support for mechanical couplers. All the electronic parts, cable routing, and battery power supply are mounted on a firm platform above the upper layer of the central body. A salient feature of this design is that all the leg modules are detachable, thus making frequent mechanical modifications and part replacement an effortless and quick process.

The highly stable design of the robot allows it to be used as a moving platform, which gives users the flexibility to mount different equipment on it like manipulators, surveillance instruments, and even military weapons. To demonstrate the same, we have already installed an in-house one degree of freedom manipulator with a gripper. Another novel feature of Prisma is its ability to turn any angle in a single step without changing its position, which is possible by the steering module. This feature aids the robot in moving smoothly in a congested labyrinth-like environment.

This section first discusses the overall dimensionality and geometry of the mechanical structure. It then shifts our focal point on its components, i.e., central body, leg module, and the mounted manipulator. Finally, we describe the actuator and sensor mounting on the whole robot.

## A. Geometry of Prisma

Prisma has a symmetric cubical design with an overall height of approximately 50 cm. It is 50 cm long, with 33 cm between the front and the hind legs, extending to 69 cm while walking as shown in Figure-3-a) & b). It has a 56 cm lateral spacing between the right and left leg planes, which only changes transitory during the turning phase. Two actuators provide the horizontal motion, which has a maximum limit of approximately 34 cm. Each leg has one actuator for its vertical motion, which gives it a maximum limit of approximately 13 cm. The robot weighs around 25 kgs with the battery packs. The static center of mass (COM) of Prisma (in a configuration similar to Figure-3-b) is approximately 22 cm above the ground.

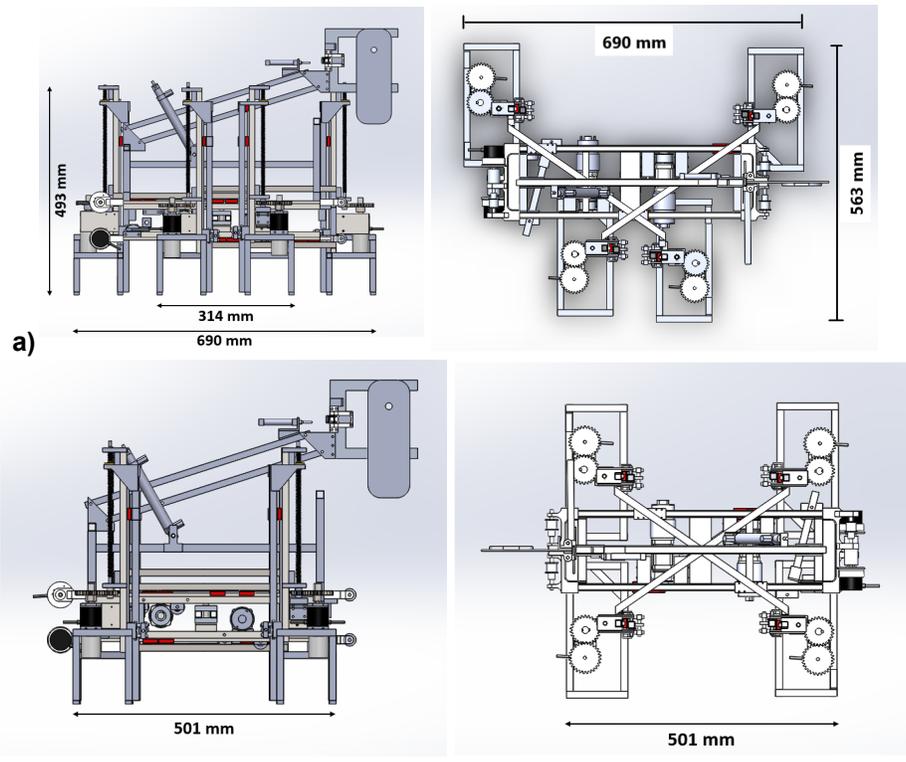

**Fig. 3** -Dimensions - a) Side view and top view in an extended state; b) Side view and top view in base state

## B. Design specifications of Prima

All the hardware specifications and actuator details are mentioned in Table-1 & 2. For better understanding, the mechanical design of Prima is explained in four different parts, which are as follows:

1. Design specifications of the Central body
2. Design specification of Sliding module
3. Design specification of Leg module
4. Design specification of the 1-DOF manipulator

**Table-I-** Actuator specifications

| Sr No | Part Name | Details | Parent Assembly |
|---|---|---|---|
| 1 | DC Motor | Banebots RS775 Motor - 18V + P6S Gearbox with 64:1 reduction | Central Body-Upper/ Lower layer |
| 2 | DC Motor | Banebots RS775 Motor - 18V + P6S Gearbox with 11.35:1 reduction | Leg Module |
| 3. | DC Motor | Cytron Worm Geared Motor (Wira) - Left - 12V | Central body- Steering Mechanism |
| 4. | Pneumatic cylinder | Air Pneumatic Cylinder, Single Rod Air Cylinder 32mm x 200mm | 1-DOF manipulator |

**Table-II**-Mechanical hardware specifications

| Sr No | Part Name | Details | Parent Assembly |
|---|---|---|---|
| 1 | Linear Slider | HIWIN MGN12C Linear Guide Rail + Block | Central Body-Upper/ Lower layer |
| 2 | Linear Slider | HIWIN MGN7C Linear Guide Rail + Block | Leg Module |
| 3. | Lead Screw | Trapezoidal 4 Start Lead Screw,304 SS, 8mm Thread, 2mm Pitch with Brass Nut | Leg Module |
| 4. | Pulley | Aluminum GT2 Timing Pulley, 26 T | Central Body-Upper/ Lower layer |

*1) Design specifications of the Central body*

The three major components of the central body are the Upper layer, the Lower layer, and the steering mechanism. The CAD model of the central body of Prisma is presented in Figure-4. A custom-made lazy susan joint connects the two similar layers to provide a rotational degree of freedom for the steering. Both the layers contain a DC motor (Banebots 64:1) and two parallel industrial slider rails which aids the horizontal translation motion of the sliding module. We have used 20-gauge SS square pipes for all the structures in the central body after comparing bending & fatigue strength in SolidWorks with available economic structural element options.

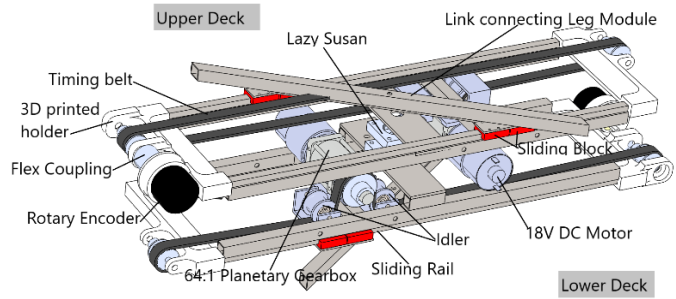

**Fig. 4**- Labelled CAD model of the central body

*1.1. Upper & Lower layer*

The upper and lower layer forms the primary central structure of the robot. The detailed CAD design of the upper & lower layer is shown in Figure-5 a) & b). An H-shaped structure is manufactured for both layers with two equal-length parallel SS pipes. An industrial slider rail (MGN) is mounted on each parallel SS pipe. Making both rails parallel was a crucial part of manufacturing, as a bit of misalignment can disrupt the horizontal motion of the sliding module on these rails. For this alignment, two 3D printed customized jigs were used. The DC motor (Banebots 64:1) for horizontal motion is mounted on another structural channel perpendicularly connecting the parallel SS square pipes.

Both the SS square pipes are now connected at the ends by 3D printed customized modules designed to support a toothed aluminium pulley, each with 25 teeth and 31 mm outer diameter. A timing belt is used to convert the rotational motion of the DC motor to the to and fro horizontal motion of the sliding module on the parallel sliding rails. An encoder is coupled to one of the pulleys for the sliding module's translational motion feedback. This translation motion of the sliding module will move the leg horizontally and thus move the Prisma forward. 3D printed idler pulleys were also used astride the motor to increase the belt's contact angle on the driver pulley (coupled to motor) and reduce slippage. The oblique link of the sliding module was used to connect the sliding blocks of the two rails, as shown in Figure-5.

The Lower layer design is an approximate mirror of the upper layer, with the difference being in the placement of DC motors and the orientation of the link connecting the sliding blocks as shown in Figure-5 b).

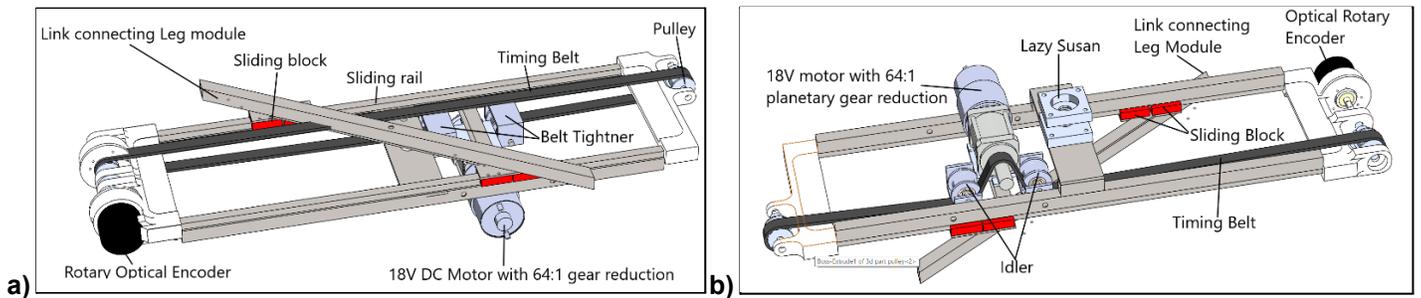

**Fig. 5**- Labelled CAD model of a) Upper layer; b) Lower layer with custom Lazy susan revolute joint

*1.2. Steering Mechanism - Lazy Susan Joint*

Lazy-Susan is a rotary joint that helps two bodies to rotate independently about a common center axis. It is sandwiched between upper and lower layers, as shown in Figure-5, and provides independent rotating aid between both the layers. Commercially available Lazy-Susan joints are costly and unsuitable for scenarios when the center of mass of the upper body is not coaxial with the joint axis. Thus, we designed a new customized lazy-susan rotation joint by coupling two bearings concentrically by a SS shaft to resolve these drawbacks, as shown in Figure-6. This customized joint also aided in keeping both the actual robot's price and weight as low as possible. An off-the-shelf worm gear motor (Cytron) with an in-built encoder was mounted on the lower layer to control the yaw motion and avoid back-drivability. A helical gear is coupled to the motor, meshed to another helical gear that is concentric to the lazy susan joint & fixed with the upper layer. Thus if we hold one layer fixed, the motor will rotate the other layer and vice versa.
.

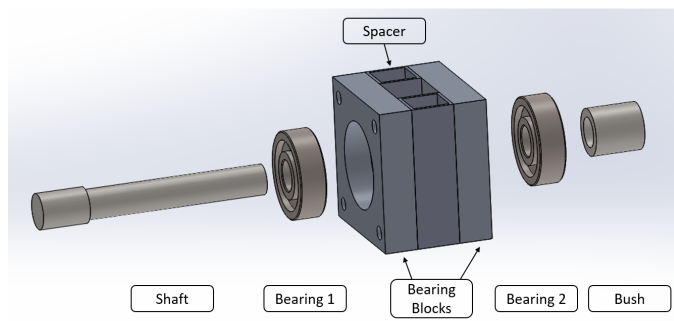

**Fig. 6**-Exploded view assembly of customized Lazy susan joint

The principle of the steering mechanism can be well understood using Figure-7. In the diagram, the legs of the robot are A, B, C, D, with one pair of diagonal legs (AC) attached to one layer (Lower) and another pair (BD) attached to another layer (Upper) via a sliding module. So for turning Prisma by an angle θ, both layers have to turn the same angle in the same direction consecutively, which is achieved by putting one diagonal pair of legs in contact with the ground (AC), supporting the robot, while the other diagonal pair remains in the air (BD). So, turning the worm gear motor θ will rotate the upper layer by θ relative to the lower layer. Now we switch the ground-contacting the diagonal pair of legs to BD, and AC will remain in the air. Then the motor is turned θ in the opposite direction, which moves the lower layer in the same direction as the upper layer by θ. Therefore using this steering mechanism, Prisma can turn any angle θ in a single step without disturbing its planar coordinates.

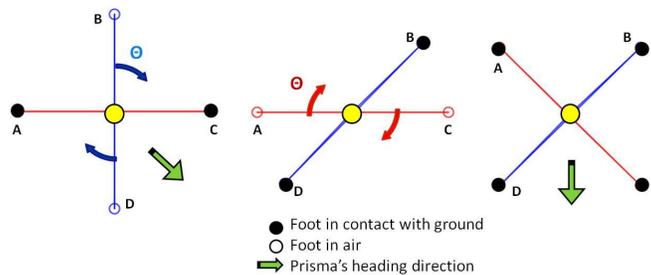

**Fig. 7** – Principle of the steering mechanism

## 2) Design specification of Sliding module

As shown in Figure-9, the sliding module consists of two diagonal leg modules welded with the help of trusses and the oblique connecting link mentioned before. Two such modules are then joined with the upper & lower layer respectively by assembling the guide blocks with the parallel sliding rails of each layer. The main challenge while designing sliding modules was to use minimum truss to keep its weight minimum, make them robust to sustain excessive fatigue and bending loads while walking and create free space for electronics equipment.

There are two possible configurations to bolt the guide blocks with the oblique link, namely X and Z, as seen from the top view. Figure-8 a) & b) shows the two configurations and their static load stress-strain simulation in SolidWorks. We can see that by applying the same load, the deformation in the X configuration is approximately 45% less than that in the Z configuration because more joints are required to make Z configuration, and the added joint play combines to give more significant bending. In addition, the extended cantilever length of the terminal of the connecting link caused considerable twisting moments on the middle bar, which caused twisting in the structure. Prisma's stride height decreases significantly due to the combination of twisting and bending in the Z configuration. Thus we chose the X configuration.

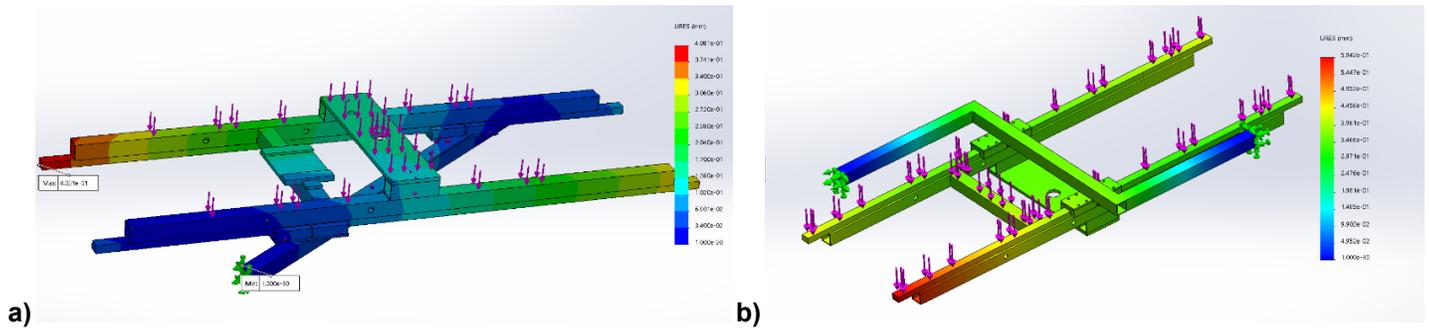

**Fig. 8**-Stress-strain analysis of a) X-shaped sliding module with maximum deformation of **0.4081 mm**. b) Z-shaped sliding module with maximum deformation of **0.5942 mm**.

## 3) Design specification of Leg Module

The leg module (Figure-9 & 10) consists of a moving foot assembly bolted to the custom nut block of a four-start lead screw of 2mm pitch for the prismatic motion. Two industrial sliding blocks (MGN 7C) are connected to the vertical SS channel of the foot at a suitable gap to ensure enough moment-bearing load. Using two blocks evenly distributes the bending moment and reduces any stride decrement due to deformation. A customized bearing-based slider is used to reduce this bending even more, as shown in Figure-10. The moving foot assembly has a rigid structure made from SS square pipes to increase each foot's contact area, thus ensuring the ZMP inside the contact polygon at every instance and thus the kinematic stability. The bottom surface of the foot structure consists of two parallel steel links, on which friction rubber pads were glued to increase the foot grip of Prisma on sloping terrain and during steering.

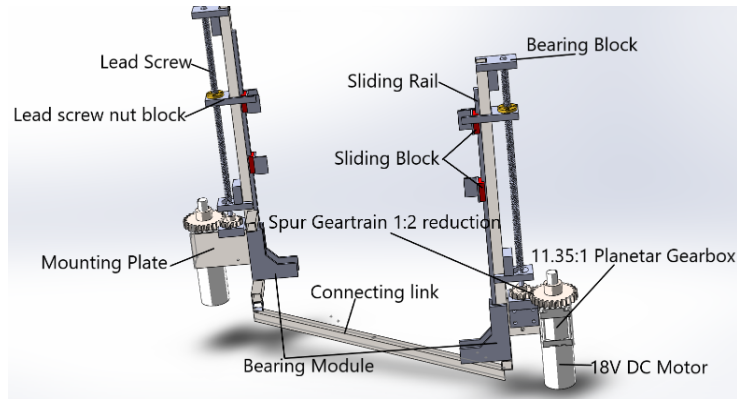

**Fig. 9-** Labelled CAD model of the sliding and leg modules

We have used a lead screw for transferring the rotary motion of the motor into linear motion, so Moment ($M_{Required}$) required to lift the Prisma is:

$$M_{Required} = W R \tan(\Phi + \alpha) \quad - (1)$$

where W is weight, R is the radius, $\Phi$ is friction angle & $\alpha$ is helix angle.

Taking weight and lead screw specification into consideration, the minimum torque required for the leg actuator is approximately **1 nm**.

An 18V DC motor (Banebots with 11.35:1 gearbox) is coupled to the lead screw with a 2:1 gear reduction to increase the rotational speed of the lead screw and thus the vertical speed of the nut. Radial bearing blocks support both ends of the lead screw to rotate while keeping its position fixed. The industrial sliding rail (MGN) and the customized bearing-based slider are mounted on the opposite face of the fixed channel parallel to the lead screw. The foot assembly is coupled to this sliding rail by the sliding blocks. Therefore, any rotational motion of the DC motor is converted to the vertical motion of the foot. This lead screw-based vertical motion of the foot has a very high mechanical advantage to support the whole robot on just two diagonal legs.

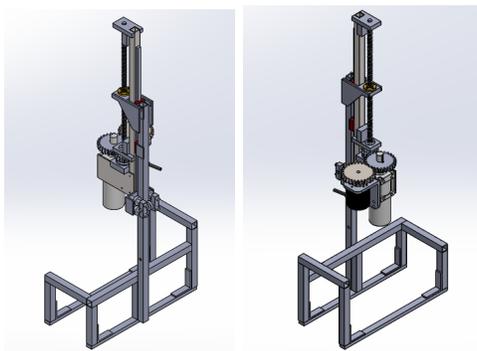

**Fig. 10 -** CAD model of leg module

### *4) Design specification of the 1-DOF manipulator*

The stable movements of Prisma allow it to act as a platform for drones & manipulators. Thus, we have also mounted an in-house manufactured 1-DOF four-bar linkage-based manipulator on Prisma. The manipulator is actuated by a pneumatic cylinder, thus giving it only two states with a height difference of approximately 1 m. A small gripper was also mounted as the end effector to grip the object and raise it to 1m in height. This manipulator is mounted to imitate one of many uses of Prisma's stable body movements.

## III. ELECTRONICS SYSTEM ARCHITECTURE

This section will describe the embedded system in detail, including sensors, motor control, communication interface, and in-house 2D-Lidar, used for perception. The primary objective is to aid Prisma in navigating autonomously in an unknown environment, which is achieved by reading and recording different sensor values. Sensor's feedback readings were used to control the movements of Prisma (Discussed in the subsequent section), making it adaptable to the environment. The electronic components which were used in Prisma are mentioned in Table-3. The sensors, motors, and other parts selected for this work are readily available off-the-shelf components that allow rapid prototyping while minimizing the total cost.

### *A. Sensory feedback*

Prisma is equipped with an IMU sensor to get the robot's orientation relative to the fixed initial position. It serves as feedback to the steering worm gear motor to turn accurately. Optical encoders mechanically coupled with all the DC motors of leg modules and central body provide the angular position and feedback for speed control of motors. Limit switch sensors are mounted at the terminal positions of the lead screws and sliding rails of the central body; to avert any off-range motion. They are also used for calibration by resetting the errors to zero. Ultrasonic distance sensors are mounted to the front legs to detect small and nearby objects like solid obstacles or ropes, which are further used to get feedback to change the trajectory of feet, giving Prisma its adaptability.

**Table-III** -Electronic hardware specification

| Equipment | Details |
|---|---|
| Optical Encoder | Orange 600 PPR |
| IMU | MPU-9150 |
| Motor Driver | Cytron 20A single channel |
| Central Platform | Raspberry pi |
| Limit Sensor | SPST Snap Action Switch |
| Distance Sensor | HC-SR04 Ultrasonic |

We have also developed an economical in-house 2D-Lidar (Figure-11) for sensing the environment, which can be further used for autonomous SLAM navigation in the future. This sensor assembly contains a high precision Laser distance sensor which is coupled to a brushed DC motor. An optical encoder is also coupled to the motor via 3D printed spur gears. Limit switches were used to reset the position of LIDAR on each swap. The Laser sensor turns a fixed angle periodically and swap-scans the surroundings to detect the position of any obstacle. The Laser sensor detects the distance of the obstacle while the encoder gives us the azimuthal angle of the obstacle with respect to the position and orientation of the robot. The rotation matrix is computed at every instance by IMU's data. This instantaneous rotation matrix is further used to get the obstacle's position with respect to the global frame. Angular resolution and rotational speed of the LIDAR can be changed at runtime. This in-house 2D-LIDAR can achieve an angular resolution of as high as 0.15 degrees. A NUCLEO-F103RB development board was also used that has STM32F103RB MCU, which is an ARM 32-bit microcontroller with a maximum clock frequency of 72 MHz and 64 MHz by internal HSI. It supports a wide range of communication protocols. We used UART communication between NUCLEO & Raspberry Pi as software addressing is not required.

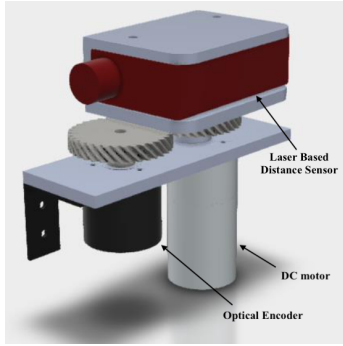

**Fig. 11-**In-house 2D Lidar assembly

## B. Motor Actuation

All the DC motors are controlled using a Cytron 20A, a pulse width modulation (PWM) based single-channel motor driver with a 12V battery back. PWM signals (of 980Hz) from the central microcontroller (i.e. Raspberry Pi) are sent to motor drivers, varying the voltage across the motor's terminals. The voltage is directly proportional to the duty cycle of the PWM signal. PID controller is used with PWM for position and velocity control of all the DC motors.

## C. Communication

We have opted for a single central MCU as shown in Figure-12 to collect data, perform computation and run control algorithms over separate MCU's because it provides better coordination and is more cost-efficient. The current design requires the central unit to support a wide range of communication protocols and a high clock frequency for proper functioning. Hence, Raspberry Pi 3 (Rpi) is selected as a central MCU for the complete functioning of the robot. IMU communicates with Rpi via I2C protocol at the frequency of 100KHz as its ACK/NACK functionality improves the error handling. The Optical encoder provides feedback to the Rpi by generating 600 pulses in a revolution. Rpi fetches data from the 2D LIDAR via UART protocol at a 115200 baud rate as software addressing is not required. Ultrasonic sensors on the front leg interact with the Rpi by two digital pins. Detailed communication flow is shown in Figure-12.

## IV. CONTROL ALGORITHM & SIMULATIONS OF PRISMA

Consider the Prisma to move on the X-Z plane (i.e ground). At any instance if the heading direction of Prisma is Φ and the coordinates of COM be ($X_c$, $Z_c$) then the planar coordinates of each foot of Prisma can be calculated using the below equation:

$$Xi = Xc \pm \frac{L}{2} * cos(\phi \pm \frac{\theta}{2}) \qquad -(2)$$
$$Zi = Zc \pm \frac{L}{2} * sin(\phi \pm \frac{\theta}{2}) \qquad -(3)$$

where ($X_i, Z_i$) is the planar coordinates of the ith foot (i = 1,2,3,4), L is the length connecting diagonal feet & Θ is the angle between front right and the front left angle at the COM.

In addition, if we consider the COM to be fixed then each foot acts as the end effector of a 3DOF RPP manipulator (Figure-13). Thus by using the D-H table we can find the transformation matrix & forward kinematics of the foot with respect to COM as shown below. Due to the low speed of Prisma, the dynamics effects are negligible. In the figure, k1, k2, k3(>k1) are dimensional constant parameters while alpha, d1 & d2 are variables that change due to the respective actuation. From the matrix, we can see that the d1 & d2 linear motions are independent and orthogonal. Thus by synchronizing the d1 & d2 with a function, the robot can walk with the chosen foot trajectory. Alpha is the turning angle of Prisma at the central lazy-susan joint used while steering.

As Prisma's leg can move independently in the orthogonal direction, multiple walking trajectories are possible by combining the vertical and horizontal motion. Different trajectories have different pros and cons, and therefore Prisma can switch between them as per the environment. We have mainly used and simulated three trajectories, i.e., Rectangular, Triangular, and Circular, as shown in Figure-14. The comparison between these trajectories based on simulation results is shown in Table-4 with the mentioned stride parameters. In the simulation, we performed a ten-step analysis of the walking of Prisma in a straight line with different trajectories on MATLAB. Notice that the stride length of the first and tenth steps are less than the other steps; this is to bring the feet back to their initial position relative to the COM. In each "Step", shown in the top part of Figure-14, the diagonal pair of the same color is in the air and moving forward using the shown trajectories while the other pair is in contact with the ground.

Maintaining the desired configuration of the trajectory is crucial as changes in the trajectory can impact Prisma's performance. So to correct the error in the trajectory, a responsive and robust control algorithm was developed. We derived the trajectory waypoints of the foot and the velocity profile for the trajectory from the MATLAB simulation. We have incorporated the following algorithm for trajectory control in the X-Y plane with respect to the global frame as shown in Figure-1. Yaw of Prisma is maintained using the Proportion-Integration (PI) controller with feedback from the IMU. We derived different trajectory functions T(Y) from the simulation, which return the Y coordinate corresponding to the X coordinate depending on the different aforementioned trajectories.

### A. Positional Control

We have a position controller that ensures the leg moves in the required trajectory and the velocity controller ensures the correct velocity is maintained. Proportional-derivative (PD) controller was used for positional control as it is highly responsive, quite stable and there is no steady-state position due to continuous motion. The Proportional-Integral-Derivative (PID) controller was used for velocity control as the trajectory has a steady-state velocity profile.

$$Distance\ moved\ by\ a\ particular\ layer\ = \frac{Total\ counts\ of\ encoder}{counts\ per\ rotation} * 2 * \pi * R \qquad - (4)$$

Where $R$ = Radius of pulley

$$Vertical\ position\ of\ Foot = \frac{Total\ counts}{Counts\ per\ turn} * Lead \qquad - (5)$$

Where *Lead* is the overall pitch of the lead screw

Error calculated for the positional control is the difference between ($X_{Required}$, $Y_{Required}$) and ($X_{Actual}$, $Y_{Actual}$), where $X_{Actual}$ (Eq.-7) and $Y_{Actual}$ (Eq-6) are the current X and Y coordinates of the foot as per the sensor data. $X_{Required}$ is the X coordinate obtained by keeping $Y_{Actual}$ in the trajectory's equation. $Y_{Required}$ is the Y coordinate obtained by keeping $X_{Actual}$ in the inverted trajectory's equation.

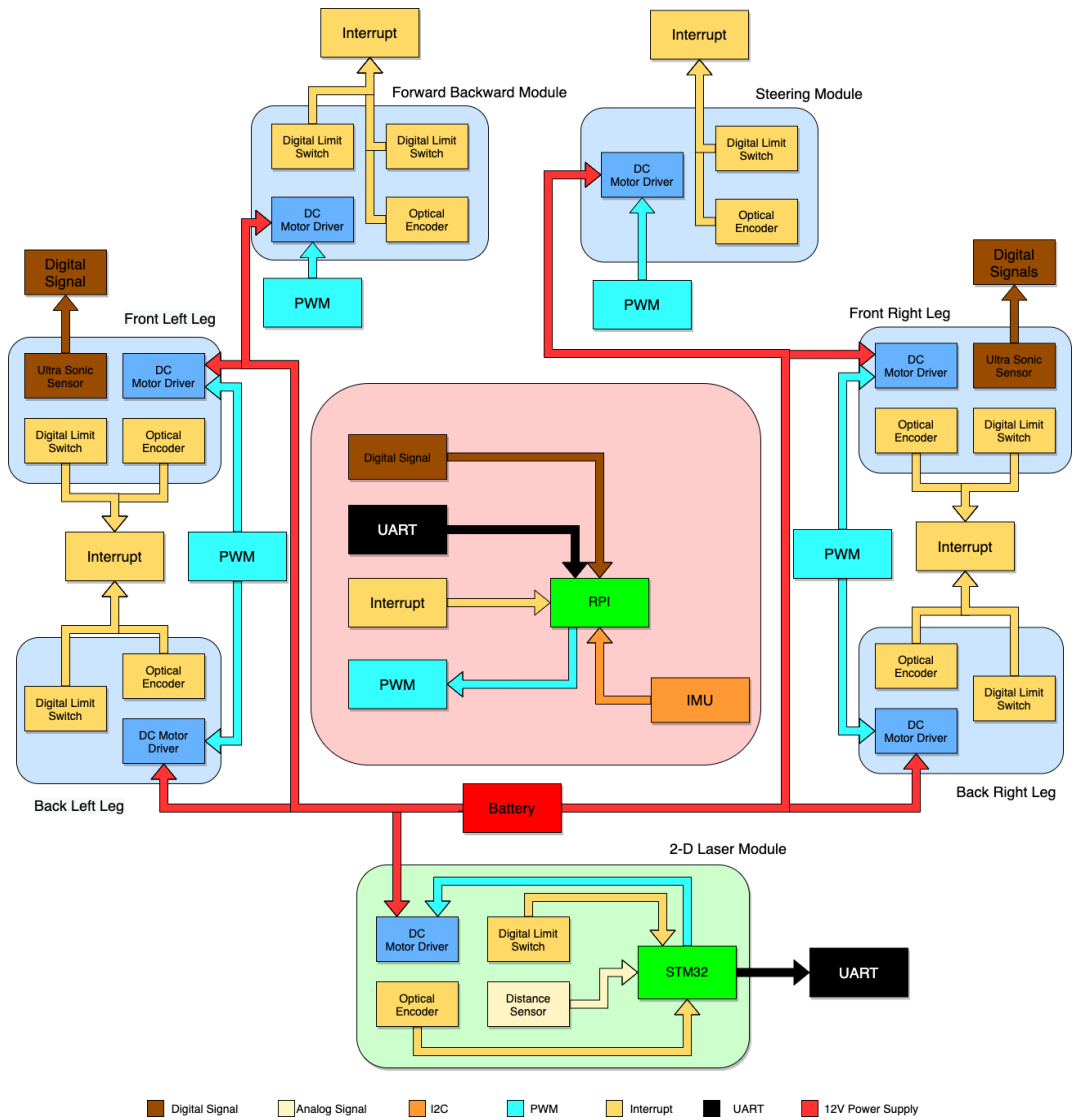

**Fig. 12**-Schematics of electronics and communication network

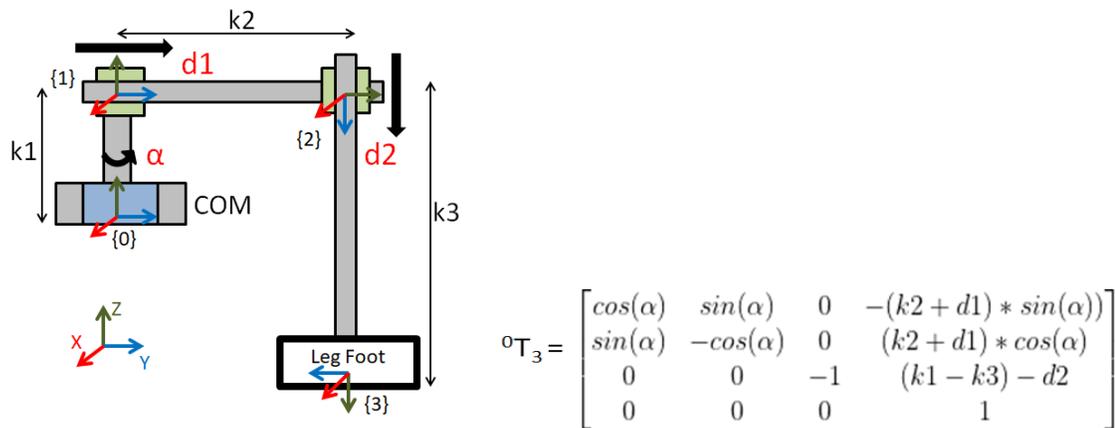

**Fig. 13**- Depiction of one leg as a 3DOF RPP manipulator & its transformation equation using DH method.

$$^0T_3 = \begin{bmatrix} cos(\alpha) & sin(\alpha) & 0 & -(k2+d1)*sin(\alpha) \\ sin(\alpha) & -cos(\alpha) & 0 & (k2+d1)*cos(\alpha) \\ 0 & 0 & -1 & (k1-k3)-d2 \\ 0 & 0 & 0 & 1 \end{bmatrix}$$

**Table-IV-** Comparison of different trajectories and the suitable surface of their use
. (L = stride length (cm), H = maximum stride height (cm))

| Trajectory | Minimum Time taken per full stride | Speed of Prisma (cm/sec) | Suitable surface |
|---|---|---|---|
| Rectangular-1 L = 34, H = 13 | 3.6 sec | 4.37 | Obstacle climbing |
| Rectangular-2 L = 34, H = 5 | 2.0 sec | 8.05 | Sandy / Muddy surface |
| Circular L = 34, H = 5 | 1.27 sec | 12.44 | Slopes |
| Triangular L = 34, H = 5 | 1.0 sec | 15.30 | Flat rigid surface |

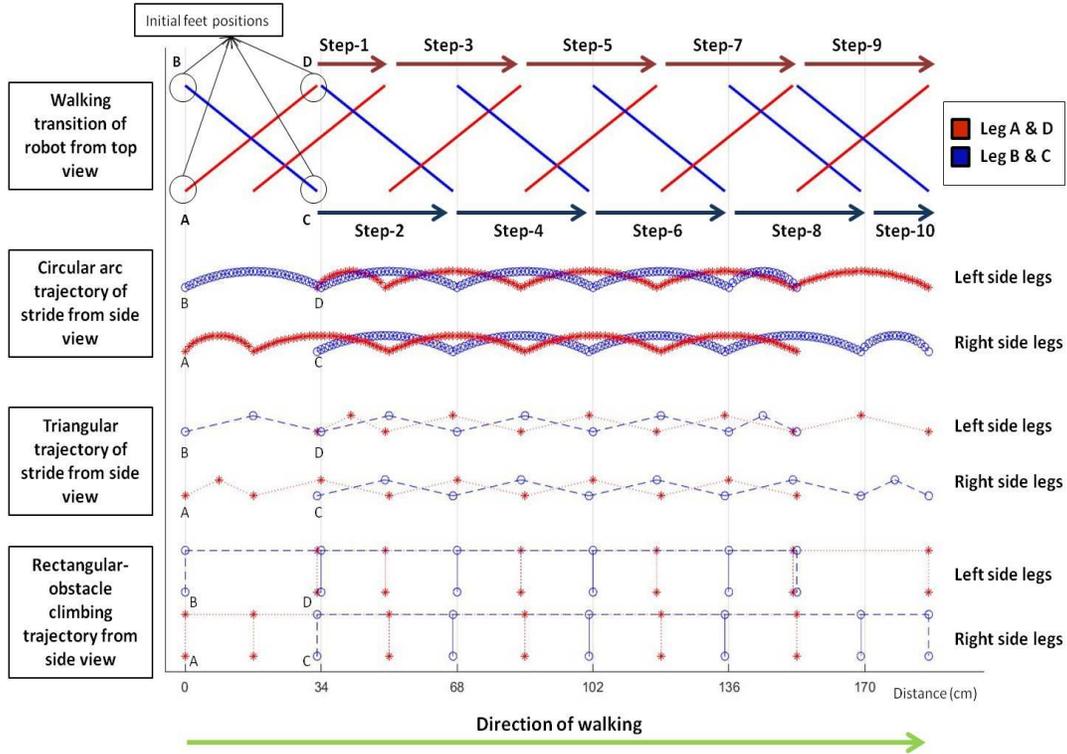

**Fig. 14** - Simulation of different trajectories of Prisma during ten steps in straight-line motion on MATLAB.

$$X_{Required} = T(Y_{Actual}) \quad - (6)$$
$$Error\_X = X_{Required} - X_{Actual} \quad - (8)$$

$$Y_{Required} = T^{-1}(X_{Actual}) \quad - (7)$$
$$Error\_Y = Y_{Required} - Y_{Actual} \quad - (9)$$

### B. Pure Pursuit velocity control

In this approach, we computed the X and Y component of the velocity using a pure pursuit-based controller [15], and then the speed controller maintains the required speed. The pure pursuit control algorithm is one of the traditional path tracking algorithms mainly used for autonomous vehicles [16], [17]. However, we have used that concept for path tracking of the foot corresponding to the required trajectory as per the MATLAB simulation result (Figure-14). A lookahead distance ($L_d$) is a fixed distance defined from the foot towards the reference trajectory in the direction of motion. The goal point is the position on the reference trajectory, $L_d$ distance away from the foot in the direction of motion. X and Y components of velocity are calculated using the goal point and current position of the foot. We have implemented a standard PID controller for quick and stable speed control in both X and Y directions. Error term in the controller is the difference between actual and calculated speed. This controller is implemented on the simple robot model in MSC Adams, as shown in Figure-15, and approximate values of the gains are computed after fine-tuning.

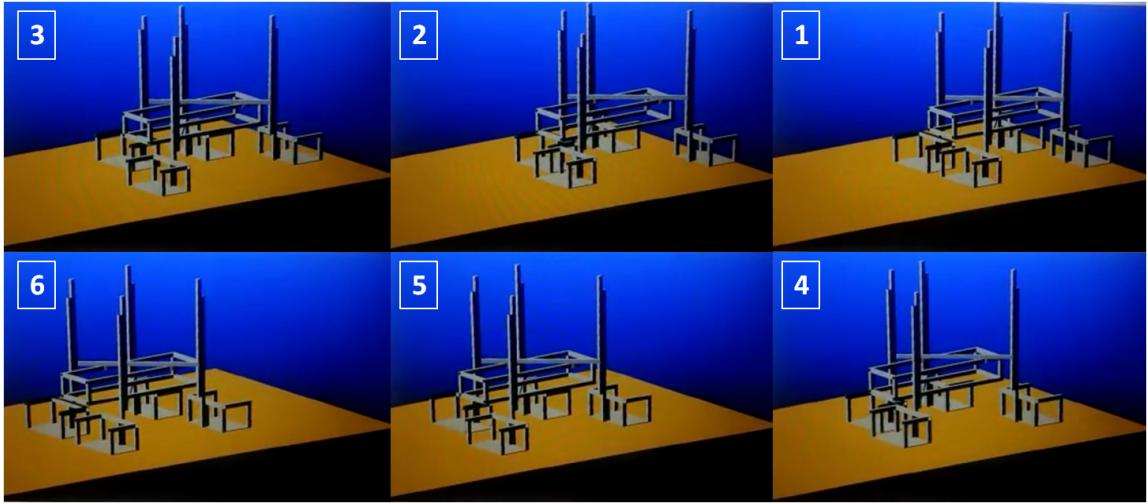

**Fig. 15 -** Motion simulation of Prisma on MSC Adams using the mentioned control algorithm and equations (1 to 6)

## V. EXPERIMENTAL RESULTS

Prisma was tested extensively in different scenarios. To verify its adaptability and smooth surmounting of hurdles, we made Prisma autonomously move through different kinds of obstacles like cuboidal wooden blocks, ropes, and ramps of different dimensions. For testing, we developed a simple wooden arena on which all the capabilities of Prisma (Walking, Turning, Obstacle detection, and trajectory variation) can be checked and improved. For actuation of the pneumatic of the manipulator, bottles are attached to store compressed air.

### A. Video Result

Video results of walking experiments of the Prisma are available on *https://youtu.be/18yQprfimvI?t=227* (3:47 onwards). Specifically, we showed Prisma doing trot gaits walking, turning, and overcoming different obstacles (blocks, ropes, and slopes). Additionally, trajectory transition can be seen for different obstacles showing Prisma's adaptability. Moreover, the dimensional, weight and geometric characteristics are explained at the start of the video.

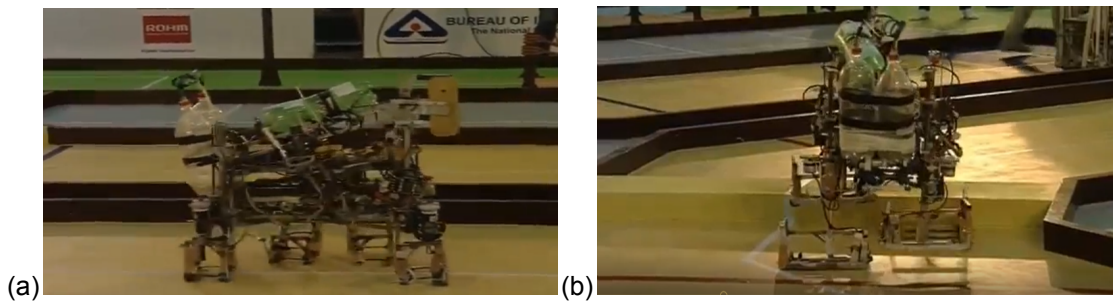

(a) (b)

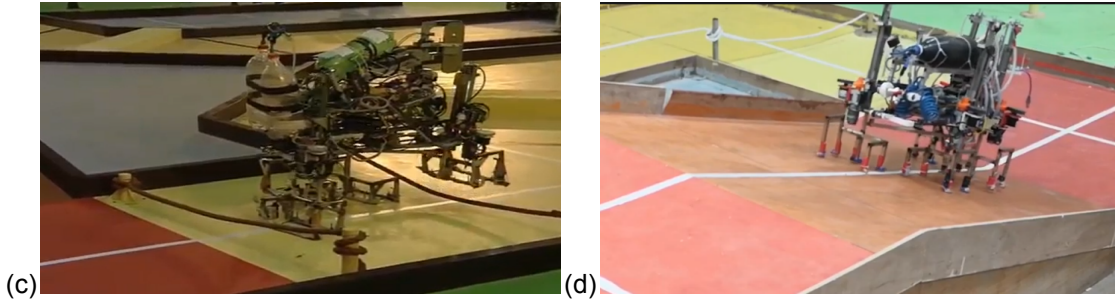

(c)       (d)

**Fig. 16-** Prisma in different scenarios. a) Walking on a flat surface, b) Obstacle climbing, c) Movement through ropes, d) Walking on the ramp.

1) Turning - During testing, Prisma effectively detects the fencing wall of the path using the in-house 2D Lidar and turns 45° accurately.

2) Cuboidal obstacle of height 10 cm - As the front ultrasound sensors detect the cuboidal obstacle, Prisma changes its foot trajectory from Triangular to Rectangular-1 and climbs the obstacle effectively. While climbing the obstacle of height less than 13 cm, Prisma's leg adjusts the vertical height to maintain the upper body's stability (Figure-16(a)). As the legs contact the top surface of the obstacle, a signal from the limit switch stops the vertical motion of that leg.

3) Slagged Rope of height 12 cm to 5 cm - Prisma smoothly moved through ropes by changing its trajectory to Rectangular-1/2 (Figure-16(b)). Entanglement from the rope is prevented by the smooth curvy foot design. We hypothesize that similar walking patterns can be used on sandy or muddy surfaces, albeit walking on these surfaces is not tested yet.

4) Ramp - Prisma is experimented with to move on sloped surfaces of different angles. After the first step on the ramp, the IMU sensor detects the body's tilt, and Prisma changes its trajectory to a tilted Circular trajectory so that the legs do not collide with the ramp before completing the full swing phase. Different angle slopes were tested and Prisma could move smoothly at a slope angle of at most 20° (Figure-16(c)).

## VI. CONCLUSION AND FUTURE WORK

This work presents a novel quadruped robot with prismatic joints along with its ingenious mechanical design, simulation, control framework, and experimental validation on the hardware. Prisma operates on a straightforward control algorithm and does not require expensive advanced sensors for its autonomous navigation compared to other existing autonomous quadruped robots. Instead, it can work effectively with traditional, economical sensors like ultrasonic distance sensors. Therefore, it requires fewer resources and costs under $500. The prismatic joints allow Prisma to walk in multiple scenarios and adapt to the situation by effortlessly changing the foot's trajectory. Additionally, the robot can turn smoothly in any direction with zero radii without changing its position. This feature allows Prisma to operate fluently in complicated industries. To show the use of Prisma as a highly stable walking platform, we have also demonstrated the use of a 1 DOF manipulator as an example. A feedback-based Pure pursuit controller was used for its autonomous navigation.

The robot's maximum speed is less compared to other agile quadruped robots, reaching a maximum forward speed of 0.2 m/s, which is one of the drawbacks of Prisma. Due to its slow linear movement, Prisma is not easy to use in a dynamic, rapidly changing environment. Another drawback is that Prisma is a bit cumbersome. Although its modular design allows it to disassemble rapidly for transport, its heavy SS structure hampers it. Future work involves increasing the movement speed of the robot and decreasing its weight significantly by using carbon fiber channels as its structure member. In addition, the incorporation of SLAM-based navigation using distance sensors & in-house 2D Lidar will be performed in future revisions.

## VII. ACKNOWLEDGEMENT

This research work was supported by the Indian Institute of Technology (IIT), Roorkee. We thank our Faculty Advisor, *Dr. Shailesh Ganpule*, Assistant Professor and Robotics faculty, *Dr. P.M. Pathak*, Professor at the Department of Mechanical and Industrial Engineering, for their insight and expertise that greatly assisted the research and significantly improved the manuscript.

--- X ---